\documentclass[sigconf]{acmart}
\renewcommand\footnotetextcopyrightpermission[1]{} % removes footnote with conference information in first column
\usepackage{booktabs} % For formal tables

\usepackage{multirow}
\usepackage{mathrsfs}
\usepackage{amsmath}
\usepackage{subfigure}

\begin{document}
\title{A Latent Feelings-aware RNN Model for User Churn Prediction with Behavioral Data}

\author{Meng Xi}
\affiliation{
  \institution{Zhejiang University}
  % \streetaddress{232 Wensan Road, 5th floor of east building}
  \city{Hangzhou}
  \state{Zhejiang}
  \country{China}
  \postcode{310007}
}
\email{ximeng@zju.edu.cn}

\author{Zhiling Luo}
\affiliation{
  \institution{Zhejiang University}
  %\streetaddress{232 Wensan Road, 5th floor of east building}
  \city{Hangzhou}
  \state{Zhejiang}
  \country{China}
  \postcode{310007}
}
\email{luozhiling@zju.edu.cn}

\author{Naibo Wang}
\affiliation{
  \institution{Zhejiang University}
  %\streetaddress{232 Wensan Road, 5th floor of east building}
  \city{Hangzhou}
  \state{Zhejiang}
  \country{China}
  \postcode{310007}
}
\email{wangnaibo@zju.edu.cn}

\author{Jianwei Yin}
\affiliation{
  \institution{Zhejiang University}
  %\streetaddress{232 Wensan Road, 5th floor of east building}
  \city{Hangzhou}
  \state{Zhejiang}
  \country{China}
  \postcode{310007}
}
\email{zjuyjw@cs.zju.edu.cn}

% \author{Anind Dey}
% \affiliation{
%   \streetaddress{Mary Gates Hall 370}
%   \institution{University of Washington}
%   \country{Canada}
% }
% \email{anind@uw.edu}

% The default list of authors is too long for headers.
% \renewcommand{\shortauthors}{B. Trovato et al.}

\begin{abstract}
  Predicting user churn and taking personalized measures to retain users is a set of common and effective practices for online game operators. However, different from the traditional user churn relevant researches that can involve demographic, economic, and behavioral data, most online games can only obtain logs of user behavior and have no access to users' latent feelings. There are mainly two challenges in this work: 1. The latent feelings, which cannot be directly observed in this work, need to be estimated and verified; 2. User churn needs to be predicted with only behavioral data. In this work, a Recurrent Neural Network(RNN) called LaFee (Latent Feeling) is proposed, which can get the users' latent feelings while predicting user churn. Besides, we proposed a method named BMM-UCP (Behavior-based Modeling Method for User Churn Prediction) to help models predict user churn with only behavioral data. The latent feelings are names as satisfaction and aspiration in this work. We designed experiments on a real dataset and the results show that our methods outperform baselines and are more suitable for long-term sequential learning. The latent feelings learned are fully discussed and proven meaningful.
  
  % In this work, a method named OSTRUC (Online Service oriented Transfer Learning method for User Churn prediction) is proposed, which can be used along with other models to solve the problem of data insufficiency. Further more, a machine learning method called LaFee (\textbf{La}tent \textbf{Fee}ling) is proposed, which can get the users' latent feelings while predicting user churn to solve the second problem. We designed experiments on a dataset of Uno to prove the effectiveness of OSTRUC and LaFee. The result shows that our method performs better than our base lines. And the latent features and are fully discussed and proven meaningful.
\end{abstract}

% 
% The code below should be generated by the tool at
% http://dl.acm.org/ccs.cfm
% Please copy and paste the code instead of the example below.
%
% \begin{CCSXML}
%   <ccs2012>
%     <concept>
%     <concept_id>10003120.10003121.10003122</concept_id>
%     <concept_desc>Human-centered computing~HCI design and evaluation methods</concept_desc>
%     <concept_significance>500</concept_significance>
%     </concept>
%     <concept>
%     <concept_id>10003120.10003121.10003122.10003332</concept_id>
%     <concept_desc>Human-centered computing~User models</concept_desc>
%     <concept_significance>300</concept_significance>
%     </concept>
%     <concept>
%     <concept_id>10003120.10003121.10003122.10003334</concept_id>
%     <concept_desc>Human-centered computing~User studies</concept_desc>
%     <concept_significance>300</concept_significance>
%     </concept>
%     <concept>
%     <concept_id>10003456.10010927</concept_id>
%     <concept_desc>Social and professional topics~User characteristics</concept_desc>
%     <concept_significance>100</concept_significance>
%     </concept>
%   </ccs2012>
% \end{CCSXML}

% \ccsdesc[500]{Human-centered computing~HCI design and evaluation methods}
% \ccsdesc[300]{Human-centered computing~User models}
% \ccsdesc[300]{Human-centered computing~User studies}
% \ccsdesc[100]{Social and professional topics~User characteristics}


\maketitle

\section{Introduction}
The term "user churn" is used in the information and communication technology industry to refer to the customer who is about to leave to find a new competitor \cite{Kasiran2014Customer, Hudaib2015Hybrid}. The problem of user churn could be different for online games and services. Because users can obtain the games directly on corresponding web pages, operators cannot flag users' churn by canceling of account or uninstalling of applications. The problem of user churn has also been taken seriously in the online game area\cite{Castro2015Churn}. In the field of online games, user churn is always judged by the logout-login interval. For instance, one could be flagged "churn" if the player has not logged in for a week. 

A traditional way to solve the problem of churn prediction is feature engineering + data labeling + prediction model training. These kinds of methods are mainly relied on the features which are already existed or well-organized in the origin data. However, the results of these methods depend too much on the qualities of the features selected from the feature engineering and cannot cope with the situations when features are difficult to get. In many cases, we can get only the users' behaviors rather than the well-organized features, leading to the failure of these methods. In the meantime, the processes of these methods were always needed to be redo when users' data changed rather than utilize the origin behavior sequences directly, i.e., these methods work with poor scalability and heavy workload. Therefore, we propose a behavior-based modeling method for user churn prediction (BMM-UCP), which reduces the churn prediction from a classification problem into a regression problem. This allows us to make full use of in-game data which could help predict users' logout time and meet the mission of user churn prediction.

% A traditional way to solve the problem of churn prediction is feature engineering + data labeling + training prediction model. However, the final result of this kind of method is dependent on the features extracted to a large extent. In addition, at the beginning of a project, because the amount of login and logout data generated is relatively small, it usually faces the problem of insufficient training data. Therefore, we propose a behavior-based modeling method for user churn prediction (BMM-UCP), which changes the churn prediction from a classification problem into a regression problem. This allows us to make full use of in-game data which could help predict users' logout time and meet the mission of user churn prediction.

In this work, in addition to being able to predict user churn, we also need to extract the latent feelings of the user through our model. Because by using these latent feelings, operators can take measures against lost users in a more timely and targeted manner. However, most researches regard machine learning methods as black boxes, which can return a result or predict something as long as the features are put in. For the end-to-end model, even the feature engineering is omitted. The only thing need to do is to throw the raw data into the "black box" and the desired model will be trained. To fit the mapping from input to output better, the models could be rather complex and deep, which result in that the meaning of the neurons of hidden layers are not important and become physically meaningless as long as the model can achieve a satisfactory loss or accuracy. In this work, besides BMM-UCP, we designed a recurrent neural network named \textit{LaFee} to estimate the users' latent feelings including \textit{satisfaction} and \textit{aspiration}. And this model can complete churn prediction at the same time. In the process, there are mainly two challenges:
\begin{enumerate}

  % \item Lack of data. The game just released for less than 3 months. Both the number of users and the number of logouts per user are small. The traditional methods like sequence analysis and deep learning method cannot work well on this kind of data set.
  % \item The information expressed by the user's logout behavior is limited, and a lot of work needs to be done during the feature engineering phase to ensure the estimated effect. In addition, one calculation can be performed only when he or her log out, which make it hard to estimate the emotional changes of the user during the game.
  \item The ground truth of users' feelings cannot be obtained directly. In this scenario, we cannot get the user's feelings directly or through physical indicators like heart rate.
  % \item the data noise is large. U-n-o is a card-based online game where users' operations, including login and logout, are susceptible to interference from other factors outside the game. For example, things like player access to the phone, network fluctuations, friends chat, etc. could have an impact on the player's behavior in the game and introduce noise into the data set.
  \item User churn needs to be predicted with only behavioral data. In the case where the user's real information such as age, occupation, etc. is missing, it could be really difficult and time-consuming to perform feature engineering on the user's behavior sequence.
\end{enumerate}

Here is the simple explanation of satisfaction and aspiration. Satisfaction indicates the pleasure a user get from every part of the game or the fulfillment degree of his or her expectations or needs. Different from satisfaction, aspiration indicates the wish or desire to perform an action. Both satisfaction and aspiration are used to express the latent feeling of a user from different aspects. For instance, after playing 1v1 or 2v2 matches for a long time, the aspiration of a player to play one more match may be decreased while the aspiration to share the game or combat gains might increase. In this process, the satisfaction will change because of the game difficulty, match mechanism or other factors. The latent feelings will be discussed in detail later.

% we believe that the hidden layers stored some latent features of the user if the model is trained well and can reach a high accuracy.

% There have been quiet a number researches that study and predict the satisfaction of a user. However, as far as we know, these works are finished with a ground truth of satisfaction, which could be a score created by the user mostly. In this work, we are committed to building a machine learning model whose hidden layers are also meaningful and explainable.

The main contributions are as follows: 
\begin{itemize}
  \item A recurrent neural network, called LaFee, is proposed, which can predict the time interval after an action of a user. This method can be used to perform churn prediction at the same time and is useful for most scenarios of games or services.
  \item We designed a behavior-based modeling method to help models predict user churn (BMM-UCP). 
  % We re-modeled the problem of user churn prediction and proposed BMM-UCP which could make full use of the data during the using process. 
  BMM-UCP can turn the classification problem of churn prediction into a regression problem and then turn it back again to make full use of the in-game data.
  \item The users' latent feelings are introduced, learned and discussed. In this work, the latent feelings are named as satisfaction and aspiration, not for that they match the real feelings of the users, but for that, they are proven own same properties as the user satisfaction and aspiration, respectively.
  % \item Hypothesises about user's latent feeling are proposed and discussed. Experiments are given to verify them as well.
  % \item We summarized some common problems that may appear in log files so that other researchers could have a reference.
\end{itemize}

\section{Related Works}
\subsection{User satisfaction}
As far as we know, most of the researches on user satisfaction are focused on the field of computer-human interaction. There are mainly two kinds of approaches to satisfaction modeling: qualitative approaches and quantitative approaches \cite{Yannakakis2008How}. In the qualitative approaches, researchers focused on modeling the incorporating flow in computer games to evaluate satisfaction or analyze the user satisfaction by clustering methods \cite{Cowley2008Toward,Lazzaro2004Why}. Since we want to be able to quantify the latent feelings of users through our model, we mainly investigate the quantitative approaches of satisfaction.

There are a lot of good researches in this field. One of the most common research lines is to propose hypothesizes or research questions, collecting interaction data, investigating user satisfaction, designing features (feature engineering), predicting satisfaction through the features and verify the thesis proposed at the beginning \cite{Klimmt2009Player,Wang2018Evaluating,Alhanai2017Predicting}. User participation is necessary for these kinds of researches. Otherwise, researchers cannot get the ground truth of the proposed prediction model or verify the validity of their features. The ground truth of satisfaction can be obtained by letting the participants score directly or analyzing their physiologic data. It is hard to be put into practice in situations like online services and online games where the users are not available. A model geared to Skype overcomes this problem with an objective source- and network-level metrics like bit rate, bit rate jitter and round-trip time to evaluate user satisfaction \cite{Chen2006Quantifying}. However, this kind of research is confined to a specific situation or area. In this work, we try to estimate user satisfaction through the state of the hidden layer in the process of predicting the time sequence of user actions. 

\subsection{Churn prediction}
Churn prediction is always a hotspot in both academia and industry.. When we need to model user churn, we need to take into account the user's use history, including state sequences, behavior sequences, and these reasons may come from external interference and accidents. 

Usually, classification algorithms are used to solve the user churn detection and prediction problem. AC Bahnsen et al. constructed a cost-sensitive customer churn prediction model framework by introducing a new financial-based approach, which enables classification algorithms to better serve business objectives \cite{Bahnsen2015A}. Z Kasiran et al. applied and compared the Elman recurrent neural network with reinforcement learning and Jordan recurrent neural network to the prediction of the loss of the subscribers of the abnormal telephones \cite{Kasiran2014Customer}. A Hudaib et al. mixed K-means algorithm, Multilayer Perceptron Artificial Neural Networks (MLP-ANN) and self-organizing maps (SOM) to establish a two-stage loss prediction model, which was tested on real datasets \cite{Hudaib2015Hybrid}.

The problem of user churn has also been taken seriously in the game area. Since the game manufacturer only got user behavior data, EG Castro et al. used four different methods to converted them into a fixed length of data array, and then took these items as input, trained probabilistic classifiers using k-nearest neighbor machine learning algorithm, which made better use of the information in the data and getting better results \cite{Castro2015Churn}.

However, in practical applications, due to the different criteria of churn, we need to train and run several models under different criteria to get satisfied results. Therefore, we proposed a model which can predict the time that may last after a user's logging out. And this time can be used to calculate various user churn rates.

\subsection{Recurrent neural network}
The recurrent neural network realizes the mapping from the input sequence to the output sequence by adding a self-looping edge in the neural network and has a wide range of applications in the problems of recognition, prediction, and generation. However, the performance on long-term dependencies is poor, and the difficulty of convergence increases with the length of the sequence \cite{Bengio1994Learning}.

\begin{figure}
  \includegraphics[width=1\columnwidth]{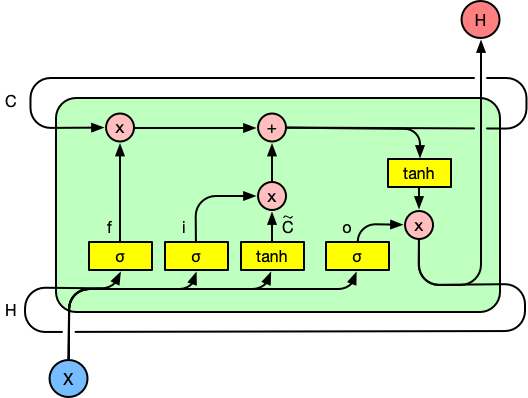}
  \caption{Sample graph of LSTM cell.}
  \label{fig:lstm}
\end{figure}

Therefore, S Hochreiter et al. proposed a long short-term memory (LSTM) algorithm to solve this problem \cite{Sepp1997Long}. The design of the LSTM neural unit is shown in Figure \ref{fig:lstm}. Its calculation formula is as follows: 

\begin{equation}
  f_t = \sigma(W_{f}\cdot[H_{t-1}, I_t] + b_{f})
\end{equation}
\begin{equation}
  i_t = \sigma(W_{i}\cdot[H_{t-1}, I_t] + b_{i})
\end{equation}
\begin{equation}
  \widetilde{C}_t = \tanh(W_{c}\cdot[H_{t-1}, I_t] + b_{c})
\end{equation}
\begin{equation}
  o_t = \sigma(W_{o}\cdot[H_{t-1}, I_t] + b_{o})
\end{equation}
\begin{equation}
  C_t = f_t * C_{t-1} + i_t * \widetilde{C}_t
\end{equation}
\begin{equation}
  H_t = o_t * \tanh(C_t)
\end{equation}

% \begin{figure}
%   \includegraphics[width=1\columnwidth]{rnn.png}
%   \caption{Sample graph of RNN cell.}
%   \label{fig:rnn}
% \end{figure}

There are many improvements and variants based on RNN or LSTM. FA Gers et al. enhanced LSTM expression and experimental results \cite{Gers2000Recurrent} by adding "peephole connections" between internal cells and multiplication gates. J Koutník et al. improved the performance and training speed of the network by reducing the number of RNN parameters, dividing the hidden layer into independent modules and inputting their respective time granularities, enabling them to perform iterative calculations at their respective clock rates. \cite{Koutn2014A}. J Bayer et al. used the advantages of variational reasoning to enhance rnn through latent variables and built Stochastic Recurrent Networks\cite{Bayer2014Learning}. N Kalchbrenner et al. extended the LSTM to apply it to multidimensional grid spaces to enhance the performance of LSTM on high-dimensional data such as images\cite{Kalchbrenner2015Grid}. Due to the extensive use of RNN and LSTM and a large number of variants, it is difficult to judge whether the RNN structure used in the current scenario is optimal. Therefore, some researchers have analyzed and evaluated RNN, LSTM and its variants in various tasks such as speech recognition and handwriting recognition to eliminate people's doubts in this area. At the same time, they found and proved that the forgetting gate and output activation function is the most The key component \cite{Jozefowicz2015An, Greff2015LSTM}.

RNN and LSTM are widely used and studied in various fields. Y Fan et al. used a recurrent neural network with two-way long-term memory cells to capture correlation or co-occurrence information between any two moments in speech, parametric text-to-speech (TTS) synthesis\cite{Fan2014TTS} . K Cho et al. encodes and decodes the symbol sequence \cite{Cho2014Learning} by combining two RNNs into one RNN codec. K Gregor et al. Constructed a deep loop attention writer for image generation by combining a novel spatial attention mechanism that simulates the concave of the human eye with a sequential variational automatic coding framework that allows iterative construction of complex images. PLD) Neural Network Structure\cite{Gregor2015DRAW}. A Ando et al. achieved a dialogue-level customer satisfaction modeling method \cite{Ando2017Hierarchical} by jointly modeling two long-term short-term regression neural networks (LSTM-RNNS).

% In summary, there are three major innovations in our work:

% \begin{itemize}
%   \item We extend the concept of user satisfaction by introducing aspiration.

% \end{itemize}

\section{Data Overview \& Preprocessing}

% \subsection{Data \& Preprocessing}

The dataset used in this work is from an online card game operated by our cooperative company. The dataset consists of log files from 89,379 users, containing 5,887,094 pieces of log data. The log files record users actions except match behaviors (the battle details in a match). Therefore, the basic rules of the game are hidden in this work, which can enhance the versatility of the method. The logs are formatted into JSON files. Each piece of log consists of three types of data:

\begin{itemize}
  \item log\_id: the identity of the log. The event that generates the log is indicated, like LoginRole, InviteLog, ConsumeItem, etc.
  \item raw\_info: the detailed information recorded in it, including login platform, IP address, server ID, etc. The information type and amount vary depending on the log\_id.
  \item timestamp: the time when the event occurred.
\end{itemize}

An example of a raw log is shown below (some private information was hidden):

\begin{quote}
  \begin{scriptsize}\begin{verbatim}
  [
    ...
    {
        "log_id": "LoginRole",
        "raw_info": {
            "account_id": "22*********",
            "ip": "4*.***.**.*81",
            ...
        },
        "timestamp": "2018-01-25 07:34:02"
    },
    ...
    {
        "log_id": "QuickMatch2V2",
        "raw_info": {
            "total1": 43,
            "total2": 43,
            "total3": -37,
            "total4": -49,
            ...
        },
        "timestamp": "2018-01-25 07:38:34"
    },
    ...
    {
        "log_id": "LogoutRole",
        "raw_info": {
            "account_id": "22*********",
            "ip": "4*.***.**.*81",
            ...
        },
        "timestamp": "2018-01-25 08:08:05"
    },
    ...
  ]
  \end{verbatim}\end{scriptsize}
\end{quote}

JSON files are appropriate for data transmission but not friendly for machine learning. The event recorded in a log could be an action taken by a user, an event triggered by the game system or a change of the user's state like gold, experience, etc. Sometimes we need to analyze the log\_id and raw\_info together to get the action user performed behind the piece of log. There are problems appeared in the log files. Some are found at the very beginning of the research, the others are discovered in the middle of the work. Some data problems may never be found and fixed if there is nothing unnatural in the experience results. There are mainly four types of log error: event name(log\_id) error, the incompleteness of paired data, duplicate records of the same event, chaotic sequence of events. Our processing method will be introduced below.

Firstly, we corrected logs with log\_id wrong. For instance, all the "PrivateGame" are recorded as "LoginRole". This problem was discovered because the "LoginRole" logs' amount are almost half more than "LogoutRole". Secondly, the logs with repeated meanings were merged together. For instance, players will be deducted 18 gold coins at the beginning of each match because of the game mechanics and generate a log with log\_id "Trade", while logs with log\_id "MatchInfo" are used to record the beginning of a match as well. Thirdly, we reordered all the logs. There are occasions that several events take place at the same time (with the same timestamp), like "LoginRole" and "DailySign", where the logs would be arranged randomly. It is unreasonable that "DailySign" appears before "LoginRole". This can lead to errors in subsequent processing. Therefore, We reordered the logs according to the proper order in which the events occurred. Finally, We inferred the users' sequences of state and time interval through the sequence of behaviors. The relationship between the three sequences is: in a time slice, the user performs an action $a$ under state $s$, and waits for the time $t$ before executing the next action. 

After all these processes, we turned logs into 29-dimensional vectors to facilitate the training of neural networks. There are 8 dimensions of states, 19 dimensions of actions and 1 dimension of the time intervals. The states and actions obtained are listed in Table \ref{tab:data}. And the relationship among state, action, and time interval in the same time slice is explained in the following section.

\begin{table}
  \caption{States and actions obtained after preprocessing.}
  \label{tab:data}
  \begin{tabular}{c|cc} 
    \toprule
    state& \multicolumn{2}{c}{action}\\
    \midrule
    Gold& LoginRole& RewardAchievement\\
    Experience& LogoutRole& InviteLog\\
    EmojisSent& ReplaceRole& ShareLog\\
    GiftsSent& PrivateGame& FollowLog\\
    AchievementGot& QuickMatch1V1& PraisePlayRound\\
    ItemNum& QuickMatch2V2& RoomModeCreate\\
    GradeUp& DailyTaskFinish& ConsumeItem\\
    OnlineDuration& DailyTaskReward& GuideInfo\\
    & DailySign& AdsLog\\
    & DailySignReward& \\
    \bottomrule
  \end{tabular}
\end{table}

\section{Our Model}
In this chapter, we will introduce LaFee and BMM-UCP in detail. As we mentioned, LaFee is designed based on BMM-UCP, so BMM-UCP will be elaborated first for better understanding.

\subsection{BMM-UCP}

BMM-UCP is a behavior-based modeling method for user churn prediction. In general, it converts the classification problem of the churn prediction into a regression problem that predicts the duration of the user's logout status based on the behavioral data. Then, the regression result would be converted back to the classification problem through different time criteria to obtain better accuracy.

\begin{figure}
  \includegraphics[width=1\columnwidth]{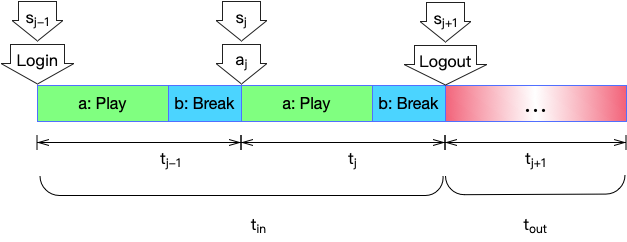}
  \caption{Illustration of Time Slice in This Work.}
  \label{fig:time}
\end{figure}

Traditionally, the churn prediction was treated as a classification problem. In the online game scenario, it's hard to tell if a user is really churned, so we need to define user churn by time criterion $\tau$, such as 1 day, 3 days and 7 days, which is widely used in industry. If a user does not log in during time $\tau$, he or she will be judged as churn. Unlike previous works, we can't get physical information about users such as their income, occupation, etc. It is really hard and time-consuming to do feature engineering with only behavioral data. In order to make full use of the existing data, we transfer the loss prediction problem into a regression problem of behavior time interval $t$ prediction. Based on the above analyses, we remodel the problem as below:

\begin{definition}
  Observing a set of game user log files. Each file can be transferred into a playing sequence $\zeta$. By transformming all log files, we can get $\mathcal{D} = \{\zeta_1 , \zeta_2 , ..., \zeta_N\}$ where $\zeta_i = \{(s_j,a_j,t_j)_{j\in{1...T}}\}$. Every $\zeta_i$ is an action sequence of a user. Within a time slice, the relation of $(s_j, a_j, t_j)$ is that player takes $a_j$ under state $s_j$ and cost time $t_j$ (see Figure \ref{fig:time}). If $a_j$ is logout, the type of $t_j$ is off-game time and belong to $t_{out}$. Otherwise, $t_j$ is in-game time and belong to $t_{in}$. Given a time $\tau$ as a user churn criterion, train a model $\mathbb{M}$ that minimizes the variance between $t_j$ (of $t_{out}$) and $t_j^* = \mathbb{M}(s_j, a_j)$, and can maximize the churn prediction accuracy (see Formula \ref{eq:accuracy}) under criterion $\tau$ at the same time.
  %generate the satisfaction and aspiration of the user at the same time.
\end{definition}

\begin{equation}
    \begin{split}
      &accuracy = \\ &\frac{\Sigma_i^{N_{t_{out}}}1((t_{outi}\ge\tau \land t_{outi}^*\ge\tau) \lor (t_{outi}<\tau \land t_{outi}^*<\tau))}{N_{t_{out}}}
    \end{split}
  \label{eq:accuracy}
\end{equation}

% And our main research tasks are as follows:
% \begin{enumerate}
%   \item We need to construct a method that can make models fit better and get better prediction results with fewer login and logout data.
%   \item The method we constructed can detect the user's latent feelings in real time, which can help game operators to personalize the retention measures to players.
%   \item For the latent feelings generated by our model, we need to prove their effectiveness through statistical analysis and data mining methods.
% \end{enumerate}

\subsection{LaFee}

With the existing data, we need to predict the churn of users. At the same time, we learned that it is inadequate to only predict whether users will lose. It is often too late to predict the loss of users, and it is impossible to understand the reasons for the loss of users. Therefore, online service operators wish to have a way, in addition to predict users' churn but also can give users real-time subjective feelings, which could help them to respond in advance.

\begin{figure}
  \includegraphics[width=1\columnwidth]{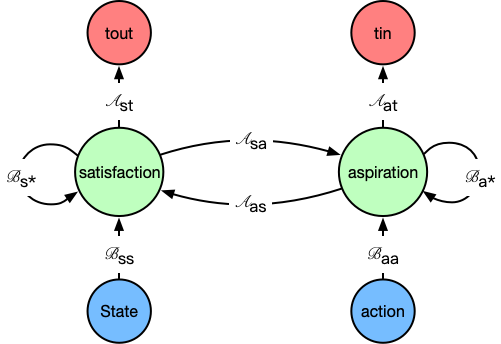}
  \caption{Graphic model of LaFee.}
  \label{fig:graphmodel}
\end{figure}

Predicting $t_{in}$ and $t_{out}$ are two relatively independent processes, as we discussed, $t_{in}$ is primarily determined by the user aspiration, and $t_{out}$ is primarily determined by the user satisfaction. So we designed two models for predicting $t_{in}$ and $t_{out}$, which are independent of each other but part of their parameters are shared (see Figure \ref{fig:graphmodel}). We take $\mathscr{A}$ as a linear translation and $\mathscr{B}$ as a non-linear translation.

LaFee consists of two processes: calculate $t_{in}$ and calculate $t_{out}$. Here we will talk about the process of calculating $t_{in}$ first. As t is $t_{in}$, satisfaction depends on the current state and previous satisfaction. Then we can calculate the aspiration of the current moment by the satisfaction we just got, the action of this moment and previous aspiration. Finally $t_{in}$ could be obtained by aspiration. Formulas are shown below.

\begin{equation}
  satisfaction_t = \mathscr{B}_{ss}(state_t) + \mathscr{B}_{s*}(satisfaction_{t-1})
  \label{eq:oval_satin}
\end{equation}
\begin{equation}
  \begin{split}
    aspiration_t = \mathscr{A}_{sa}(satisfaction_{t}) + \mathscr{B}_{aa}(action_t) \\ + \mathscr{B}_{a*}(aspiration_{t-1})
  \end{split}
  \label{eq:oval_aspin}
\end{equation}
\begin{equation}
  t_{in} = \mathscr{A}_{at}(aspiration_t)
  \label{eq:oval_in}
\end{equation}

The case where t is $t_{out}$ is shown by Formula \ref{eq:oval_aspout}, \ref{eq:oval_satout}, \ref{eq:oval_tout}. And it is clear that $\mathscr{B}_{ss}, \mathscr{B}_{s*}, \mathscr{B}_{aa}, \mathscr{B}_{a*}$ are shared by two processes. The rest translations are owned by each.

\begin{equation}
  aspiration_t = \mathscr{B}_{aa}(action_t) + \mathscr{B}_{a*}(aspiration_{t-1})
  \label{eq:oval_aspout}
\end{equation}
\begin{equation}
  \begin{split}
    satisfaction_t = \mathscr{A}_{as}(aspiration_{t}) + \mathscr{B}_{ss}(state_t) \\ + \mathscr{B}_{s*}(satisfaction_{t-1})
  \end{split}
  \label{eq:oval_satout}
\end{equation}
\begin{equation}
  t_{out} = \mathscr{A}_{st}(satisfaction_t)
  \label{eq:oval_tout}
\end{equation}

% \subsection{Calculate Time Interval}

\begin{figure}
  \includegraphics[width=1\columnwidth]{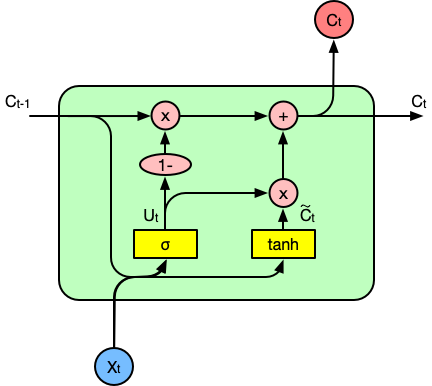}
  \caption{Sample graph of LaFee cell.}
  \label{fig:lafee}
\end{figure}

As we have discussed above, both satisfaction and aspiration have sequential dependencies. Therefore, we need to redesign the method of RNN to establish the time sequence relationship between the value of satisfaction or aspiration. We built a simple variant of LSTM to calculate and store the values of satisfaction and aspiration. We hid the output layer, merged the cell state and the hidden layer, and simplified the design of the input gate and the forget gate (see Figure \ref{fig:lafee}).

The first step is to calculate satisfaction. We calculate $Us_t$ by formula \ref{eq:13} using sigmoid function as activation function, current state and previous satisfaction as input. As $Us_t$ indicates what to remember, $1-Us_t$ indicates what to forget. Then we calculate a candidate satisfaction $\widetilde{Cs}_t$ that used $\tanh$ function and same input as $Us_t$ (formula \ref{eq:14}). Finally, new satisfaction is an addition of the values to be remembered (calculated by multipling $Us_t$ and $\widetilde{Cs}_t$) and memories left by previous satisfaction (gained by multipling $1-Us_t$ and $sat_{t-1}$). The formulas \ref{eq:13} - \ref{eq:15} correspond to formula \ref{eq:oval_satin}. 
% The satisfaction of a player is depend on the state sequence and previous satisfaction. 

\begin{equation}
  Us_t = \sigma(W_{us}\cdot[sat_{t-1}, state_t] + b_{us})
  \label{eq:13}
\end{equation}
\begin{equation}
  \widetilde{Cs}_t = \tanh(W_{cs}\cdot[sat_{t-1}, state_t] + b_{cs})
  \label{eq:14}
\end{equation}
\begin{equation}
  sat_t = (1 - Us_t) * sat_{t-1} + Us_t * \widetilde{Cs}_t
  \label{eq:15}
\end{equation}

The second step is to calculate the aspiration. Firstly, aspiration depends on the action the player has taken recently. For instance, one may get tired and lose some aspiration if dozens of matches were played in the past few minutes. Secondly, the previous aspiration can affect that of this time slice because aspiration is consecutive to some extent. Thirdly, current satisfaction also has an impact on the aspiration. The calculation of aspiration is basically the same as satisfaction except the influence of satisfaction is introduced in the last step. Detailed steps are shown in formulas \ref{eq:16} - \ref{eq:18}, which correspond to formula \ref{eq:oval_aspin}. 

\begin{equation}
  Ua_t = \sigma(W_{ua}\cdot[asp_{t-1}, action_t] + b_{ua})
  \label{eq:16}
\end{equation}
\begin{equation}
  \widetilde{Ca}_t = \tanh(W_{ca}\cdot[asp_{t-1}, action_t] + b_{ca})
  \label{eq:17}
\end{equation}
\begin{equation}
  asp_t = (1 - Ua_t) * asp_{t-1} + Ua_t * \widetilde{Ca}_t + W_{sa} \cdot sat_t + b_{sa}
  \label{eq:18}
\end{equation}

The time interval of $t_{in}$ is consisted of two parts. As shown in Figure \ref{fig:time}, a player executes action $a$(not logout) under state $s$, the corresponding $t_{in}$ is mainly determined by the two parts, i.e., $t_{in}^a$(play part) and $t_{in}^b$(relax part). $t_{in}^a$ is mainly determined by the type of the action and the game mechanism, while $t_{in}^b$ is affected by aspiration. The final step is to get the $t_{in}$ using formula \ref{eq:19}, which corresponds to formula \ref{eq:oval_in}.

\begin{equation}
  t_{in} = W_{at} \cdot asp_t + b_{at}
  \label{eq:19}
\end{equation}

% For some instantaneous actions, the time interval only depends on the wait time. Wait time is the time between the execution of the current action and the start of the next action. For some continuous actions, the time interval also needs to involve the time consumed by the action itself, which was called basic time here.

% Firstly, there is a basic time strongly associated with the action. This basic time is consumed by the action like match time and finger movement time. Then, the wait time between two actions should depend on the player's aspiration. The aspiration corresponding to the current action should also have an impact on the wait time. For instance, you may need to take a break after a match, but not if you just click and receive a reward.

For the calculation of $t_{out}$. As we have designed, the process of calculating $t_{in}$ and the process of calculating $t_{out}$ are relatively symmetrical. The calculation process of $t_{out}$ can be clearly seen and understood through formulas below.

\begin{table*}
  \caption{Comparison among Different Approaches}
  \label{tab:comparison}
  \resizebox{\textwidth}{!}{ 
  \begin{tabular}{c|ccc|ccc|ccc|ccc|ccc}
    \toprule
    \multirow{2}{*}{Method}& \multicolumn{3}{c}{High Winning-rate}& \multicolumn{3}{c}{Low Winning-rate}& \multicolumn{3}{c}{Battle Player}& \multicolumn{3}{c}{Social Player}& \multicolumn{3}{c}{ALL}\\
    & 1 day& 3 days& 7 days& 1 day& 3 days& 7 days& 1 day& 3 days& 7 days& 1 day& 3 days& 7 days& 1 day& 3 days& 7 days\\
    \midrule
    RF& \textbf{99.17} & \textbf{96.57} & 86.27 & \textbf{99.82} & \textbf{98.91} & 91.30 & \textbf{99.37} & \textbf{96.54} & 82.08 & \textbf{99.62} & \textbf{98.64} & 88.44 & \textbf{99.56} & \textbf{97.82} & 86.78 \\
    RF\_bu& 74.09 & 81.11 & 82.32 & 82.83 & 86.87 & 87.54 & 72.03 & 81.67 & \textbf{83.60} & 80.47 & 86.69 & \textbf{87.57} & 81.18 & 89.46 & \textbf{90.87} \\
    RF\_lf& 81.10 & 86.58 & \textbf{87.81} & 88.36 & 92.81 & \textbf{93.49} & 72.40 & 82.14 & \textbf{84.42} & 85.67 & 92.24 & \textbf{93.43} & 84.75 & 92.58 & \textbf{93.49} \\ 
    \midrule
    % SVM& 98.52 & 94.58 & 79.31 & 99.82 & 99.28 & 92.75 & 99.40 & 95.18 & 79.52 & 99.52 & 98.64 & 89.80 & 99.60 & 97.90 & 87.49 \\ 
    % SVM\_bu& 5.93 & 1.03 & 0.26 & 6.73 & 0.67 & 0.15 & 18.01 & 3.22 & 0.64 & 13.29 & 1.86 & 0.23 & 12.44 & 2.15 & 0.45 \\ 
    % SVM\_lf& 4.96 & 0.78 & 0.26 & 7.92 & 0.71 & 0.23 & 16.78 & 3.36 & 0.67 & 11.61 & 1.18 & 0.24 & 11.69 & 1.82 & 0.38 \\ 
    % \midrule
    GDBT& \textbf{98.52} & \textbf{94.58} & 79.31 & \textbf{99.82} & \textbf{99.28} & 92.75 & \textbf{99.40} & 95.18 & 79.52 & \textbf{98.29} & 98.64 & 89.80 & \textbf{99.60} & 97.90 & 87.49 \\ 
    GDBT\_bu& 74.32 & 93.92 & \textbf{97.97} & 89.18 & 98.80 & \textbf{99.80} & 82.37 & \textbf{96.13} & \textbf{98.49} & 90.53 & 98.52 & \textbf{99.77} & 87.56 & 97.85 & \textbf{99.55} \\ 
    GDBT\_lf& 75.69 & 94.44 & \textbf{98.61} & 90.15 & 99.18 & \textbf{99.90} & 83.30 & \textbf{96.48} & \textbf{99.12} & 91.04 & \textbf{98.81} & \textbf{99.96} & 88.31 & \textbf{98.18} & \textbf{99.62} \\ 
    \midrule
    DNN& \textbf{98.52} & 94.58 & 76.85 & \textbf{99.64} & \textbf{98.55} & 92.39 & \textbf{98.80} & \textbf{95.18} & 75.30 & \textbf{99.24} & \textbf{98.30} & 88.78 & \textbf{99.25} & \textbf{97.70} & 86.89 \\ 
    DNN\_bu& 83.81 & 91.38 & \textbf{92.59} & 91.25 & 96.63 & \textbf{97.31} & 77.63 & 90.54 & \textbf{92.90} & 85.53 & 94.03 & \textbf{95.60} & 85.45 & 97.11 & \textbf{99.02} \\ 
    DNN\_lf& 87.44 & \textbf{95.10} & \textbf{96.63} & 93.15 & 97.60 & \textbf{98.29} & 80.22 & 91.65 & \textbf{93.85} & 87.38 & 94.82 & \textbf{95.79} & 86.57 & 97.16 & \textbf{99.13} \\ 
    \midrule
    % RNN& 87.79 & 97.38 & 99.42 & 88.80 & 98.74 & 99.89 & 81.52 & 95.53 & 98.83 & 86.61 & 98.38 & 99.77 & 87.06 & 97.82 & 99.59 \\ 
    % RNN\_bu& 82.32 & 95.36 & 98.55 & 86.85 & 97.41 & 99.68 & 78.78 & 95.03 & 98.47 & 82.13 & 97.17 & 99.55 & 80.67 & 95.73 & 99.04 \\ 
    % RNN\_lf& 95.01 & 99.21 & 99.74 & 90.15 & 99.18 & 99.90 & 83.30 & 96.48 & 99.12 & 88.51 & 98.63 & 99.86 & 88.31 & 98.18 & 99.62 \\ 
    % \midrule
    LSTM& 85.90 & 97.09 & 99.42 & 87.92 & 98.78 & 99.86 & 80.64 & 94.44 & 98.60 & 85.66 & 97.82 & 99.61 & 86.85 & 97.78 & 99.59 \\ 
    LSTM\_bu& 82.32 & 95.36 & 98.55 & 86.85 & 97.41 & 99.68 & 78.78 & \textbf{95.03} & 98.47 & 82.13 & 97.17 & 99.55 & 80.67 & 95.73 & 99.04 \\ 
    LSTM\_lf& \textbf{95.01} & \textbf{99.21} & \textbf{99.74} & \textbf{90.15} & \textbf{99.18} & \textbf{99.90} & \textbf{83.30} & \textbf{96.48} & \textbf{99.12} & \textbf{88.51} & \textbf{98.63} & \textbf{99.86} & \textbf{88.31} & \textbf{98.18} & \textbf{99.62} \\ 
    % \midrule
    % LaFee& 82.32 & 95.36 & 98.55 & 86.85 & 97.41 & 99.68 & 78.78 & 95.03 & 98.47 & 82.13 & 97.17 & 99.55 & 80.67 & 95.73 & 99.04 \\ 
    \bottomrule
  \end{tabular}}
\end{table*}

Formula \ref{eq:20}-\ref{eq:22} are used to calculate aspiration first corresponding to formula \ref{eq:oval_aspout}. The next step is to use the aspiration just obtained, combined with the state and the previous satisfaction to calculate the current satisfaction (formula \ref{eq:23}-\ref{eq:25}, corresponding to \ref{eq:oval_satout}). And $t_{out}$ is calculated at the end by formula \ref{eq:26} (corresponding to \ref{eq:oval_tout}). Detailed description will not be repeated.

\begin{equation}
  Ua_t = \sigma(W_{ua}\cdot[asp_{t-1}, action_t] + b_{ua})
  \label{eq:20}
\end{equation}
\begin{equation}
  \widetilde{Ca}_t = \tanh(W_{ca}\cdot[asp_{t-1}, action_t] + b_{ca})
  \label{eq:21}
\end{equation}
\begin{equation}
  asp_t = (1 - Ua_t) * asp_{t-1} + Ua_t * \widetilde{Ca}_t
  \label{eq:22}
\end{equation}
\begin{equation}
  Us_t = \sigma(W_{us}\cdot[sat_{t-1}, state_t] + b_{us})
  \label{eq:23}
\end{equation}
\begin{equation}
  \widetilde{Cs}_t = \tanh(W_{cs}\cdot[sat_{t-1}, state_t] + b_{cs})
  \label{eq:24}
\end{equation}
\begin{equation}
  sat_t = (1 - Us_t) * sat_{t-1} + Us_t * \widetilde{Cs}_t + W_{as} \cdot asp_t + b_{as}
  \label{eq:25}
\end{equation}
\begin{equation}
  t_{out} = W_{st} \cdot sat_t + b_{st}
  \label{eq:26}
\end{equation}

Whether calculating $t_{in}$ or $t_{out}$, our optimization direction is the same, which is to minimize variance between time intervals predicted and the real ones (see \ref{eq:loss}).

\begin{equation}
  loss = \frac{\Sigma_{j=1}^n (t_j - t_j^*)^2}{n}
  \label{eq:loss}
\end{equation}

\section{Experiments}

%  On the other hand, through data mining and statistical methods, we further discuss the role of verifying the two meanings.
% In this experiment, we compared the final performance of our model and other models the same dataset. 
% The second part is about the verification experiments of satisfaction and aspiration for model output. Through some statistical data and analysis methods, we verify the true meaning of these data. We added them as features to training to verify their effectiveness.

% Our experiments are divided into three parts. The first part is about the experiment of the effect of the model itself. 
Our experiment is mainly divided into three parts. The first part is used to validate the effectiveness of BMM-UCP and LaFee. The second part and the third part are used to analyze and verify the physical meaning and effectiveness of satisfaction and aspiration through statistical analysis and data mining methods.

Before the experiment started, we extracted 4 groups of users. First, we sorted the data by users' match winning percentage, dividing the highest 20\% into the high winning-rate group, and the lowest 20\% into the low winning-rate group. The winning percentages of the two groups were 62.66\% and 19.33\%, respectively. In addition, according to the different proportions of users' behaviors, we regard the 20\% user with the highest proportion of match behaviors as the battle player, and the user with the least 20\% of the match behaviors as the social player. The match class behaviors accounted for 18.47\% and 5.39\%, respectively.

\subsection{Performance Evaluation}

In this part, we want to study the effectiveness of our model. We apply data sets to 4 baseline methods. There are two off-the-shelf classifiers, including Random Forest (RF) \cite{liaw2002classification} and Gradient Boosting Decision Tree (GBDT) \cite{Friedman2002Stochastic}. Both of them are implemented by scikit-learn and the number of trees of RF is set as 500. The other two are deep learning approaches, including Long Short Term Memory(LSTM) \cite{Sepp1997Long}, Deep Neural Network(DNN) \cite{hinton2012deep}. They are implemented by tensorflow and keras, respectively. The time step of LSTM is sixteen and the DNN has six hidden layers.

At the same time, we apply the BMM-UCP to all the baselines. We use the suffix of \_bu to identify methods involving BMM-UCP, such as dnn\_bu. In addition, we also introduce the satisfaction and aspiration calculated by the LaFee model as input to each baseline to prove LaFee's validity. Similarly, we distinguish these methods by a suffix of \_lf. Baseline\_lf and baseline\_bu just differ on input.

All programs are executed on the same computer with 32G RAM, 2.8GHz CPU, and NVidia GeForce TitanX GPU. 70\% of data are randomly extracted and used for training and the rest are used as test set. It was ensured that all approaches are compared on the same test set. We repeated 10 times for each experiment to get the average accuracy.

In Table \ref{tab:comparison}, the prediction accuracies of the four groups and all users under three different criteria are accurate to two decimal places. For RF, GDBT, and DNN, the accuracies decrease as the increase of $\tau$, while the method involving BMM-UCP and LaFee are the opposite. This is because when $\tau$ is one day, the so-called churn has a strong randomness. In this case, most of the behavioral data are noise. BMM-UCP has no advantage. With the increase of $\tau$, the regularity of user churn and the correlation with in-game behavior are greatly improved. So our method can be better fitted. In fact, users who do not log in for 7 consecutive days are most likely to be the users who are actually lost. Under criterion of 7 days, the models with LaFee involved show the highest accuracies.

As for LSTM, the accuracies increases as the $\tau$ increases, because LSTM has introduced timing information. However, we observed that the result of introducing only BMM-UCP was worse than LSTM, while the result of introducing LaFee was better than LSTM. This is because BMM-UCP introduced the in-game data when it was involved, which means that the latent feelings within and out the game are learned together, eventually leading to overfitting of the model. LaFee learned satisfaction and aspiration semi-separately. And the performances of LSTM\_lf are always the best, which show that LaFee does learn the latent feelings of users. For all the methods and data groups, LaFee involved methods' results are better than those of BMM-UCP, which also proves the effectiveness of the satisfaction and aspiration we have learned.

There are also some interesting phenomena appeared in table \ref{tab:comparison}. For example, the accuracies of the low-winning-rate group are higher than those of the high-winning-rate group, and the accuracies of the social group are higher than those of the battle group. These will be discussed in conjunction with the figures in subsequent experiments.

% After communicating with industry experts, we learned that the standard time of loss time $\tau$ is usually 1 day, 3 days and 7 days.

\subsection{Satisfaction Analysis}

\begin{figure}
  \includegraphics[width=\columnwidth]{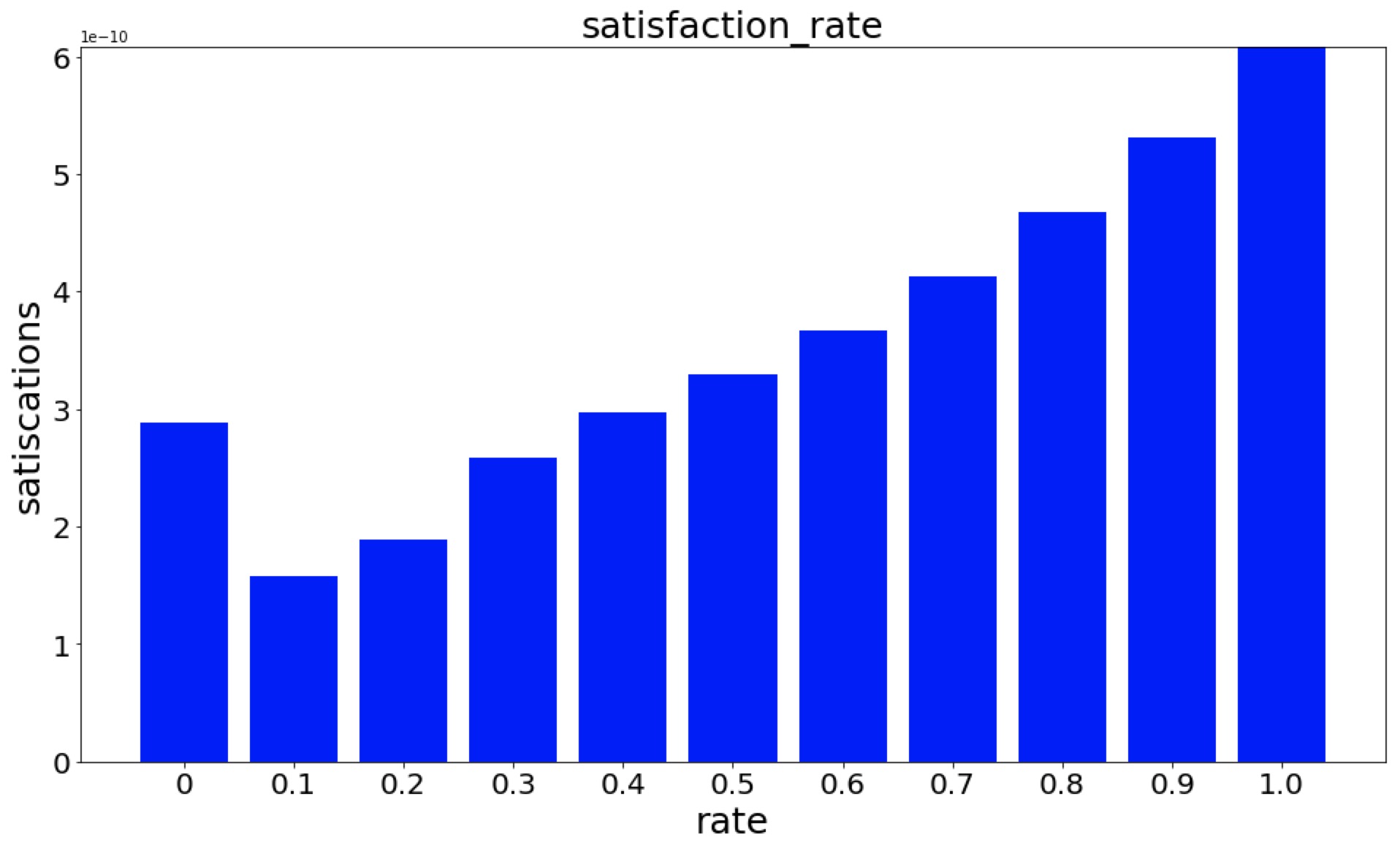}
  \caption{Satisfaction and winning rate correlation statistics.}
  \label{fig:sat_win}
\end{figure}

We first studied the relationship between user satisfaction and match winning rate (see Figure \ref{fig:sat_win}). As we all know, user satisfaction is often more affected by recent game experiences, so here we only calculate the winning percentage of the user's last 10 match games. The x-axis represents different winning rate and the Y-axis is the average of satisfaction under the corresponding winning rate. The satisfaction we find here is only a relative quantity, and its absolute value has no specific meaning. So for ease of presentation and reader understanding, we have added an offset of $1*10^{-10}$ on the Y-axis. 

It is easy to observe that in the last 10 games, except for the total defeat, the user satisfaction increased with the increase of the winning rate. Different from what we have known in the past, it might be the best choice to control the user match winning rate to around 50\%, \textit{the user satisfaction can be improved with the short-term victory.} When the user loses 10 games in a row, the satisfaction is similar to the case when the winning percentage is 40\%. This shows that \textit{compared to the losing streak, occasionally winning one or two games makes the user feel more dissatisfied and painful.}

\begin{figure}
  \includegraphics[width=\columnwidth]{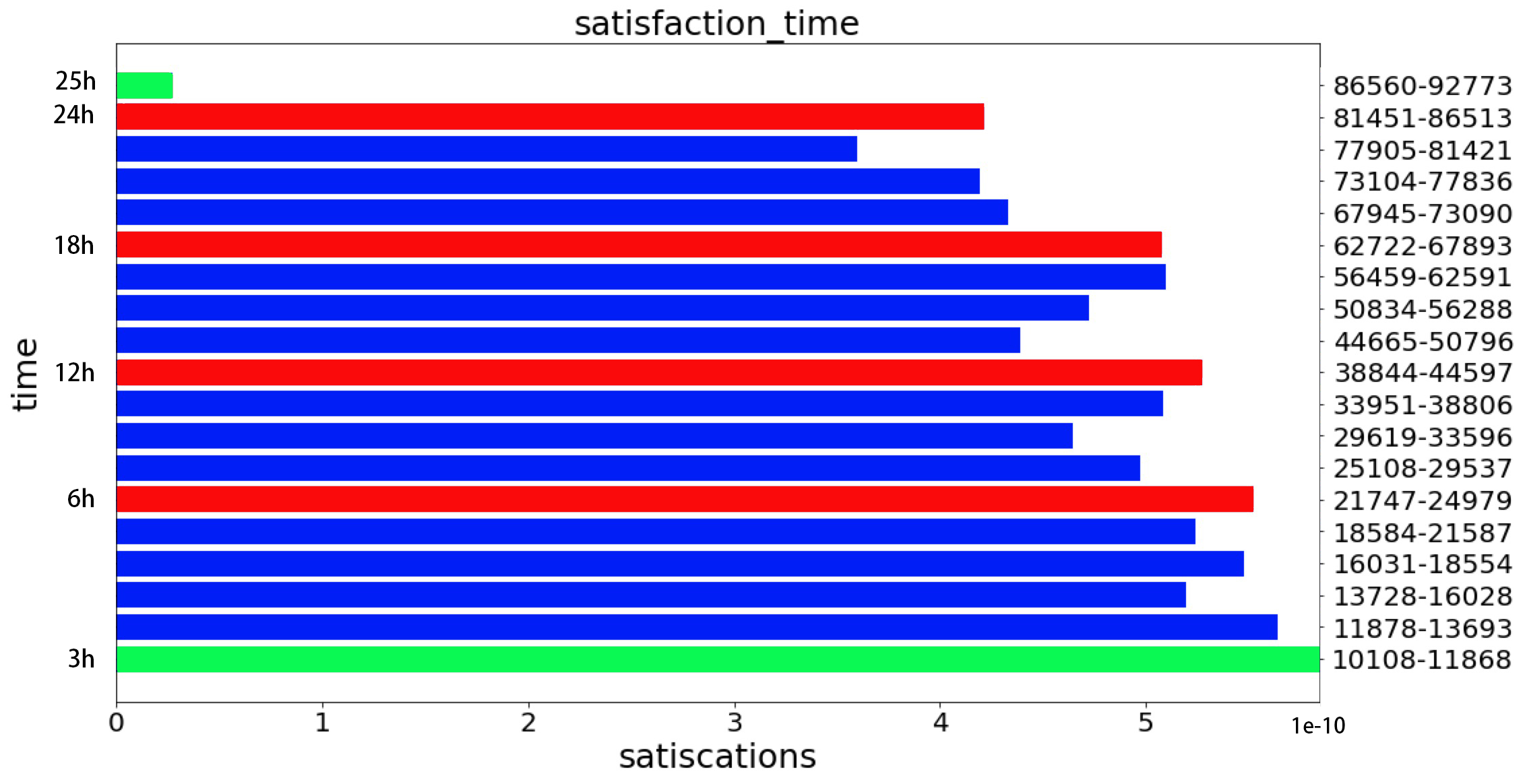}
  \caption{Satisfaction and logout time correlation statistics.}
  \label{fig:sat_time}
\end{figure}

In addition, figure \ref{fig:sat_time} can also explain why in the table \ref{tab:comparison} the prediction accuracy of the low-winning-rate group is always higher than the high-winning-rate group. This is because the satisfaction of the low-winning-rate group fluctuates more.

Then we studied the relationship between user satisfaction and user logout time (see Figure \ref{fig:sat_time}). The x-axis is the average of satisfaction in a time domain and Y axis is the corresponding logout time ($t_{out}$) domain whose unit is second. Since the value range of $t_{out}$ is very large, in order to be able to perform a detailed analysis, we conducted our analysis of that under the condition that the login time was within one day. As in the previous part, we added an offset of $1*10^{-9}$ on the X-axis for better presentation and understanding.

First, let's look at the green bar with the highest satisfaction. The corresponding time interval is about 3 hours. In other words, \textit{the user re-login once every three hours or so means that the user is satisfied with the game.} This is understandable in combination with our daily experience. 3 hours is just the effective working time for most people in the morning, afternoon or evening. Whenever the user finishes their work, they will remember to log in the game, and it does indicate that they are satisfied with the game and willing to play again and again.

Then we look at the red bar. The time intervals corresponding to the four red bars from bottom to top are about 6, 12, 18, and 24 hours, respectively. As the value of $t_{out}$ increases, the average user satisfaction wavily decreases. This shows that \textit{when users are not satisfied with the game, they tend to log out for a longer period of time.} And 6, 12, 18, 24 hours and the 3 hours analyzed are just on the peak. That is to say, \textit{the user's logout time is more consistent with the objective time interval, which means that the user is more satisfied with the game.} We also observed that when the logout time reached over 86,400 seconds (over one day), the user satisfaction would fall off sharply. In other words, \textit{one day can indeed be used as a criterion for a user's estimation of game satisfaction and the possibility of loss.} Moreover, \textit{if the value of satisfaction is equal or lower than that of 1-day-average, the user may not return in short term.}

\subsection{Aspiration Analysis}

\begin{figure*}
  \subfigure[]{\includegraphics[width=2.1in]{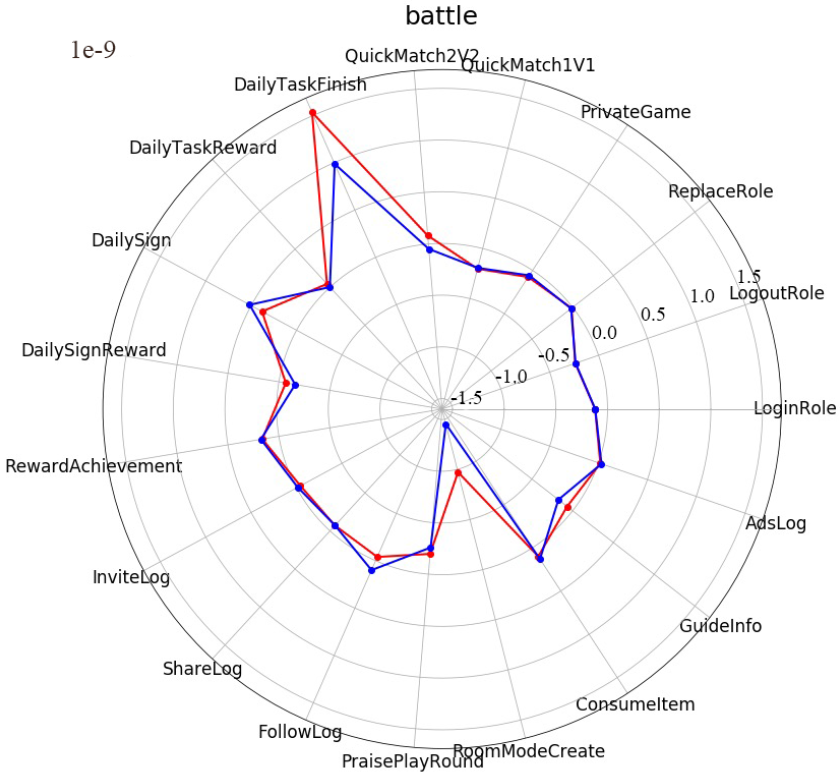}}
  \subfigure[]{\includegraphics[width=3.7in]{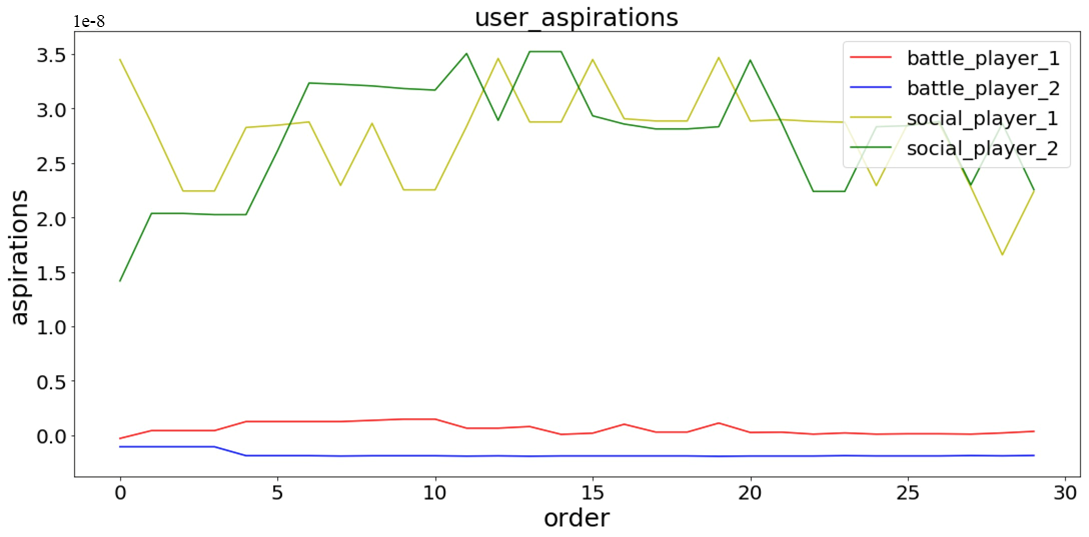}}
  \subfigure[]{\includegraphics[width=2.1in]{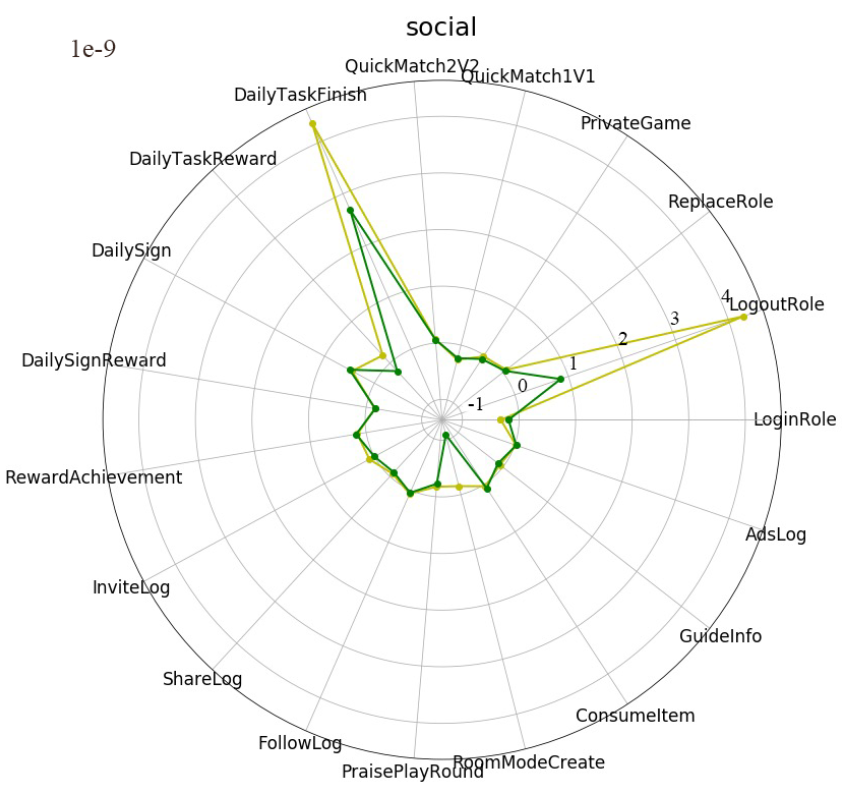}}
  \subfigure[]{\includegraphics[width=3.7in]{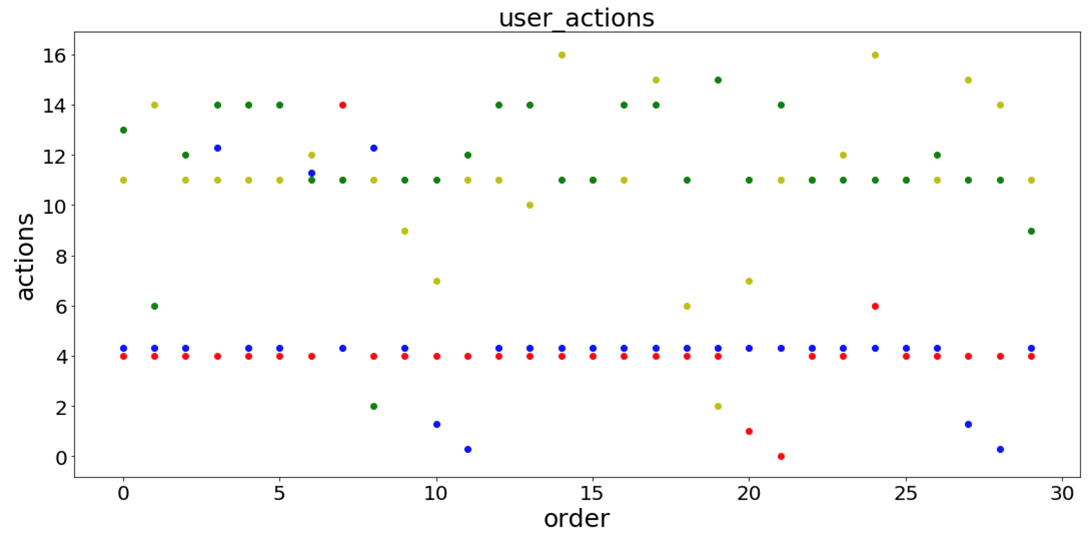}}
  \caption{Comparison among users on aspiration.}
  \label{fig:asp_action}
\end{figure*}

\begin{figure}
  \includegraphics[width=\columnwidth]{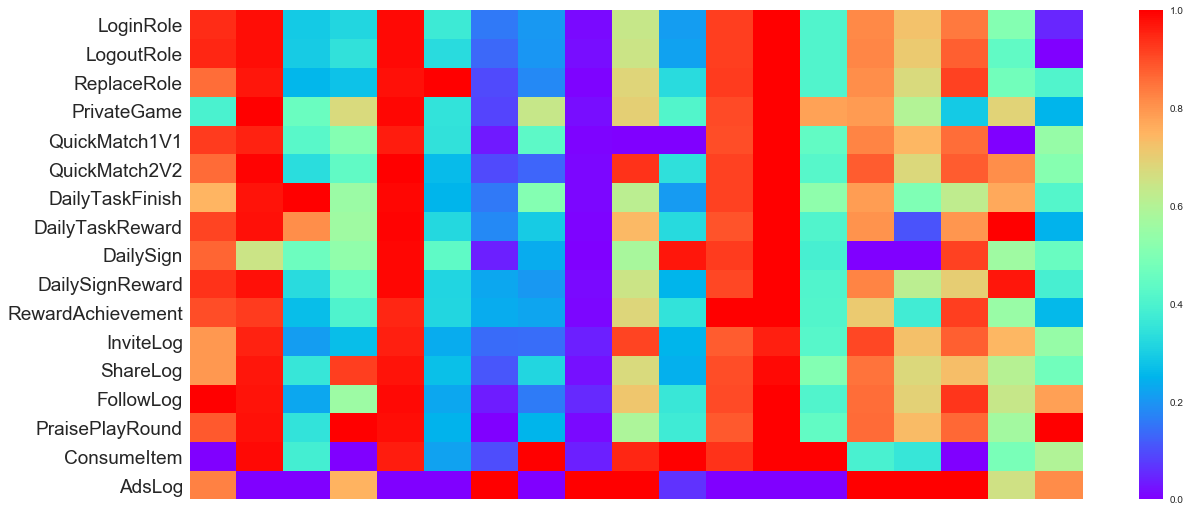}
  \caption{Colormap of average aspiration to each action.}
  \label{fig:asp_color}
\end{figure}

We extracted the aspiration corresponding to each behavior on the training set, group them according to the corresponding behavior and average them within the group. Thus we can get a 19-dimensional average aspiration vector for each type of behavior. For all 19 types of behavior, we can form a matrix of average aspiration vectors. In order to achieve a better visualization, we normalize the matrix by min-max and then show it through the colormap (see Figure \ref{fig:asp_color}).

% \begin{figure}
%   \subfigure[]{\includegraphics[width=\columnwidth]{asp-color1.png}}
%   \subfigure[]{\includegraphics[width=\columnwidth]{asp-color2.png}}
%   \caption{Colormap of average aspiration to each action.}
%   \label{fig:asp_color}
% \end{figure}

From Figure \ref{fig:asp_color} we can see that the aspiration of different behaviors is obviously different. Service operators can speculate on what users might do next through their current aspirations. 

Then we conducted a further analysis of aspiration. We randomly extracted four users, named BP1, BP2, SP1, and SP2, equally from the battle players and the social players. In Figure \ref{fig:asp_action}, All red lines or points are of BP1. Blue, yellow and green are of BP2, SP1, and SP2, respectively. Through our model, we extracted the aspiration of all the behaviors of the four users. We averaged each user aspiration, get the user's average aspiration and make two radar maps (see Figure \ref{fig:asp_action} (a)\&(c)).

It can be clearly seen that BP1 and BP2 are very similar in the case that the two users are of the same type (see Figure \ref{fig:asp_action} (a)). This situation also exists between SP1 and SP2 (see Figure \ref{fig:asp_action} (c)). However, BP1/BP2 is very different from the radar map of SP1/SP2 who are not battle players. In other words, \textit{users with similar behavioral habits will have similar aspiration radar charts.} Second, in the case where the sequences randomly taken by BP1 and BP2 are similar (mostly QuickMatch1V1, shown as 4 in Y-axis), their aspiration sequences are also very similar. The SP1 and SP2 with different sequences are taken out, although they belong to the social player group, their aspiration performances are totally different. In summary, \textit{By observing and analyzing the distribution and trend of aspiration, one can estimate the behavior distribution and action strategy of a user.}

Also, in the table \ref{tab:comparison}, the social group is always easier to predict than the battle group. As you can see from figure \ref{fig:asp_action} (b), the aspiration of the social group is more abundant. In contrast, the users of the battle group actually own not too high aspiration. They might just play for killing time.

\section{Discussion}
In this chapter, we mainly discuss a possible problematic point that does not affect the logic and understanding of the main content and the application value of our proposed method and model.

The problematic point is that why satisfaction and aspiration could be obtained by predicting the time sequence through state and action sequence. Below are the reasons:
\begin{enumerate}
  \item $t_{in}$ is mainly determined by the user's current aspiration. As shown in Figure \ref{fig:time}, a player executes action $a_1$(not log out) under state $s_1$, the corresponding $t_1$ is mainly determined by the two parts of $t_{1a}$(play part) and $t_{1b}$(relax part). $t_{1a}$ is mainly related to the type of behavior itself, such as the time of reward-taking behavior may be shorter and a game behavior may be longer. $t_{1b}$ is mainly determined by the user's current aspiration to play, when the aspiration to play is strong, the user may quickly proceed to the next action, resulting in $t_{1b}$ is very short, and vice versa.
  \item User satisfaction is mainly stored in user states and reflected by the length of $t_{out}$. Incompetence, immersion, flow, tension, challenge, etc. are usually used to express satisfaction \cite{Phan2016The}. Ultimately, these factors will be converted into data stored in the player's state sequence. It often leads to the infinite extension of the $t_{out}$ which could be the ultimate churn, i.e., a user is not satisfied with the game.
  \item User satisfaction and game aspiration will interact with each other and have a sequential relationship. When the game satisfaction is high, the players' aspiration to play is easier to maintain at a higher level. Conversely, when the game satisfaction is relatively low, players will lose their aspiration to play faster. At the same time, no matter satisfaction or aspiration, the value of the present time slice depends on the last one.
\end{enumerate}

The valadities of these three reasons have been proved throughout the article. And a simple example can also achieve the same effect. When a player plays a game, his satisfaction with the game may increase or decrease. But after a certain length of time in the game, he will certainly withdraw, that is, the aspiration to play the results of the declines. In the case of more satisfied with the game, after a period of off-line rest, the aspiration for the game will be rebounded, leading to the user logs in again for the next round of the game. In general, the off-line break time may be longer as the decrease of game satisfaction. More extreme condition is that the game is very unsatisfactory, making user completely lose the aspiration to play this game and never log on.

Our method and model are of great value to both academia and industry. On one hand, BMM-UCP can not only remodel the churn prediction problem, but also help solve the problem of underfitting of the model using only the behavior sequences, rather than introducing a new data set. User churn is a problem for almost all network games, but most games even do not accumulate to the amount that can effectively analyze the causes of user churn using traditional methods. Our approach helps developers and operators improve their games better and faster by making full use of the existing data in the system.

On the other hand, our LaFee model can output the user satisfaction and aspiration in real time while predicting user churn. Game operators can adjust the games' parameters by the performance of satisfactions, such as the correlation between the match winning-rate and satisfaction mentioned above, to provide players with better game experience. Secondly, the service operators can also use satisfaction to better estimate the user's next possible logout time. Finally, the real-time feedback from satisfaction can help service operators better understand where user satisfaction rises or falls and analyze them. Aspiration can help to segment the user groups and quickly classify users in the condition of the absence of data. At the same time, according to the real-time changes of aspiration, the operators can estimate the user's next actions and then provide more targeted service recommendations and advertisement pop-up services.

% Time interval prediction is meaningful because this can help the operator find proper timing to insert an advertisement. For instance, an advertisement could be rather annoying if the user just finished an action and want to continue to perform another immediately while it could be the opposite if the user is trying to take a break.

\section{Conclusions}
In this work, we propose a RNN model called LaFee to get latent feelings while predicting user churn, which mitigates the challenge of lacking users' real feelings. In the mean time, we introduce a method called BMM-UCP to help models predict user churn when it needs to be completed with only behavioral data. Then we carry out statistical quantitative analysis of the satisfaction and aspiration while expounding and proving the physical meaning of them. Finally, the application scenarios and meanings of our model and the latent feelings are discussed, they are proven useful for both academia and industry.

In the future, we will improve our model to predict the latent feelings more than satisfaction and aspiration, and expand the application scope of the model to get more interesting and meaningful results.

%\end{document}  % This is where a 'short' article might terminate

\begin{acks}
  Thanks

\end{acks}

\bibliographystyle{ACM-Reference-Format}
\bibliography{churn,satisfaction,rnn}

%%% -*-BibTeX-*-
%%% Do NOT edit. File created by BibTeX with style
%%% ACM-Reference-Format-Journals [18-Jan-2012].

\begin{thebibliography}{27}

%%% ====================================================================
%%% NOTE TO THE USER: you can override these defaults by providing
%%% customized versions of any of these macros before the \bibliography
%%% command.  Each of them MUST provide its own final punctuation,
%%% except for \shownote{}, \showDOI{}, and \showURL{}.  The latter two
%%% do not use final punctuation, in order to avoid confusing it with
%%% the Web address.
%%%
%%% To suppress output of a particular field, define its macro to expand
%%% to an empty string, or better, \unskip, like this:
%%%
%%% \newcommand{\showDOI}[1]{\unskip}   % LaTeX syntax
%%%
%%% \def \showDOI #1{\unskip}           % plain TeX syntax
%%%
%%% ====================================================================

\ifx \showCODEN    \undefined \def \showCODEN     #1{\unskip}     \fi
\ifx \showDOI      \undefined \def \showDOI       #1{#1}\fi
\ifx \showISBNx    \undefined \def \showISBNx     #1{\unskip}     \fi
\ifx \showISBNxiii \undefined \def \showISBNxiii  #1{\unskip}     \fi
\ifx \showISSN     \undefined \def \showISSN      #1{\unskip}     \fi
\ifx \showLCCN     \undefined \def \showLCCN      #1{\unskip}     \fi
\ifx \shownote     \undefined \def \shownote      #1{#1}          \fi
\ifx \showarticletitle \undefined \def \showarticletitle #1{#1}   \fi
\ifx \showURL      \undefined \def \showURL       {\relax}        \fi
% The following commands are used for tagged output and should be
% invisible to TeX
\providecommand\bibfield[2]{#2}
\providecommand\bibinfo[2]{#2}
\providecommand\natexlab[1]{#1}
\providecommand\showeprint[2][]{arXiv:#2}

\bibitem[\protect\citeauthoryear{AlHanai and Ghassemi}{AlHanai and
  Ghassemi}{2017}]%
        {Alhanai2017Predicting}
\bibfield{author}{\bibinfo{person}{Tuka~Waddah AlHanai} {and}
  \bibinfo{person}{Mohammad~Mahdi Ghassemi}.} \bibinfo{year}{2017}\natexlab{}.
\newblock \showarticletitle{Predicting Latent Narrative Mood Using Audio and
  Physiologic Data.}. In \bibinfo{booktitle}{\emph{AAAI}}.
  \bibinfo{pages}{948--954}.
\newblock


\bibitem[\protect\citeauthoryear{Ando, Masumura, Kamiyama, Kobashikawa, and
  Aono}{Ando et~al\mbox{.}}{2017}]%
        {Ando2017Hierarchical}
\bibfield{author}{\bibinfo{person}{Atsushi Ando}, \bibinfo{person}{Ryo
  Masumura}, \bibinfo{person}{Hosana Kamiyama}, \bibinfo{person}{Satoshi
  Kobashikawa}, {and} \bibinfo{person}{Yushi Aono}.}
  \bibinfo{year}{2017}\natexlab{}.
\newblock \showarticletitle{Hierarchical LSTMs with Joint Learning for
  Estimating Customer Satisfaction from Contact Center Calls}. In
  \bibinfo{booktitle}{\emph{INTERSPEECH}}. \bibinfo{pages}{1716--1720}.
\newblock


\bibitem[\protect\citeauthoryear{Bahnsen, Aouada, and Ottersten}{Bahnsen
  et~al\mbox{.}}{2015}]%
        {Bahnsen2015A}
\bibfield{author}{\bibinfo{person}{Alejandro~Correa Bahnsen},
  \bibinfo{person}{Djamila Aouada}, {and} \bibinfo{person}{Bj{\"o}rn
  Ottersten}.} \bibinfo{year}{2015}\natexlab{}.
\newblock \showarticletitle{A novel cost-sensitive framework for customer churn
  predictive modeling}.
\newblock \bibinfo{journal}{\emph{Decision Analytics}} \bibinfo{volume}{2},
  \bibinfo{number}{1} (\bibinfo{year}{2015}), \bibinfo{pages}{5}.
\newblock


\bibitem[\protect\citeauthoryear{Bayer and Osendorfer}{Bayer and
  Osendorfer}{2014}]%
        {Bayer2014Learning}
\bibfield{author}{\bibinfo{person}{Justin Bayer} {and}
  \bibinfo{person}{Christian Osendorfer}.} \bibinfo{year}{2014}\natexlab{}.
\newblock \showarticletitle{Learning stochastic recurrent networks}.
\newblock \bibinfo{journal}{\emph{arXiv preprint arXiv:1411.7610}}
  (\bibinfo{year}{2014}).
\newblock


\bibitem[\protect\citeauthoryear{Bengio, Simard, and Frasconi}{Bengio
  et~al\mbox{.}}{1994}]%
        {Bengio1994Learning}
\bibfield{author}{\bibinfo{person}{Y Bengio}, \bibinfo{person}{P Simard}, {and}
  \bibinfo{person}{P Frasconi}.} \bibinfo{year}{1994}\natexlab{}.
\newblock \showarticletitle{Learning long-term dependencies with gradient
  descent is difficult.}
\newblock \bibinfo{journal}{\emph{IEEE Trans Neural Netw}} \bibinfo{volume}{5},
  \bibinfo{number}{2} (\bibinfo{year}{1994}), \bibinfo{pages}{157--166}.
\newblock


\bibitem[\protect\citeauthoryear{Castro and Tsuzuki}{Castro and
  Tsuzuki}{2015}]%
        {Castro2015Churn}
\bibfield{author}{\bibinfo{person}{Emiliano~G. Castro} {and}
  \bibinfo{person}{Marcos S.~G. Tsuzuki}.} \bibinfo{year}{2015}\natexlab{}.
\newblock \showarticletitle{Churn Prediction in Online Games Using Players'
  Login Records: A Frequency Analysis Approach}.
\newblock \bibinfo{journal}{\emph{IEEE Transactions on Computational
  Intelligence \& Ai in Games}} \bibinfo{volume}{7}, \bibinfo{number}{3}
  (\bibinfo{year}{2015}), \bibinfo{pages}{255--265}.
\newblock


\bibitem[\protect\citeauthoryear{Chen, Huang, Huang, and Lei}{Chen
  et~al\mbox{.}}{2006}]%
        {Chen2006Quantifying}
\bibfield{author}{\bibinfo{person}{Kuan~Ta Chen}, \bibinfo{person}{Chun~Ying
  Huang}, \bibinfo{person}{Polly Huang}, {and} \bibinfo{person}{Chin~Laung
  Lei}.} \bibinfo{year}{2006}\natexlab{}.
\newblock \showarticletitle{Quantifying Skype user satisfaction}. In
  \bibinfo{booktitle}{\emph{Conference on Applications, Technologies,
  Architectures, and Protocols for Computer Communications}}.
  \bibinfo{pages}{399--410}.
\newblock


\bibitem[\protect\citeauthoryear{Cho, Merrienboer, Gulcehre, Bahdanau,
  Bougares, Schwenk, and Bengio}{Cho et~al\mbox{.}}{2014}]%
        {Cho2014Learning}
\bibfield{author}{\bibinfo{person}{Kyunghyun Cho}, \bibinfo{person}{Bart~Van
  Merrienboer}, \bibinfo{person}{Caglar Gulcehre}, \bibinfo{person}{Dzmitry
  Bahdanau}, \bibinfo{person}{Fethi Bougares}, \bibinfo{person}{Holger
  Schwenk}, {and} \bibinfo{person}{Yoshua Bengio}.}
  \bibinfo{year}{2014}\natexlab{}.
\newblock \showarticletitle{Learning Phrase Representations using RNN
  Encoder-Decoder for Statistical Machine Translation}.
\newblock \bibinfo{journal}{\emph{Computer Science}} (\bibinfo{year}{2014}).
\newblock


\bibitem[\protect\citeauthoryear{Cowley, Charles, Black, and Hickey}{Cowley
  et~al\mbox{.}}{2008}]%
        {Cowley2008Toward}
\bibfield{author}{\bibinfo{person}{Ben Cowley}, \bibinfo{person}{Darryl
  Charles}, \bibinfo{person}{Michaela Black}, {and} \bibinfo{person}{Ray
  Hickey}.} \bibinfo{year}{2008}\natexlab{}.
\newblock \showarticletitle{Toward an understanding of flow in video games}.
\newblock \bibinfo{journal}{\emph{Computers in Entertainment (CIE)}}
  \bibinfo{volume}{6}, \bibinfo{number}{2} (\bibinfo{year}{2008}),
  \bibinfo{pages}{20}.
\newblock


\bibitem[\protect\citeauthoryear{Fan, Qian, Xie, and Soong}{Fan
  et~al\mbox{.}}{2014}]%
        {Fan2014TTS}
\bibfield{author}{\bibinfo{person}{Yuchen Fan}, \bibinfo{person}{Yao Qian},
  \bibinfo{person}{Feng-Long Xie}, {and} \bibinfo{person}{Frank~K Soong}.}
  \bibinfo{year}{2014}\natexlab{}.
\newblock \showarticletitle{TTS synthesis with bidirectional LSTM based
  recurrent neural networks}. In \bibinfo{booktitle}{\emph{Fifteenth Annual
  Conference of the International Speech Communication Association}}.
\newblock


\bibitem[\protect\citeauthoryear{Friedman}{Friedman}{2002}]%
        {Friedman2002Stochastic}
\bibfield{author}{\bibinfo{person}{Jerome~H. Friedman}.}
  \bibinfo{year}{2002}\natexlab{}.
\newblock \showarticletitle{Stochastic Gradient Boosting}.
\newblock \bibinfo{journal}{\emph{Computational Statistics \& Data Analysis}}
  \bibinfo{volume}{38}, \bibinfo{number}{4} (\bibinfo{year}{2002}),
  \bibinfo{pages}{367--378}.
\newblock


\bibitem[\protect\citeauthoryear{Gers and Schmidhuber}{Gers and
  Schmidhuber}{2000}]%
        {Gers2000Recurrent}
\bibfield{author}{\bibinfo{person}{Felix~A Gers} {and}
  \bibinfo{person}{J{\"u}rgen Schmidhuber}.} \bibinfo{year}{2000}\natexlab{}.
\newblock \showarticletitle{Recurrent nets that time and count}. In
  \bibinfo{booktitle}{\emph{Proceedings of the IEEE-INNS-ENNS International
  Joint Conference on Neural Networks. IJCNN 2000. Neural Computing: New
  Challenges and Perspectives for the New Millennium}},
  Vol.~\bibinfo{volume}{3}. IEEE, \bibinfo{pages}{189--194}.
\newblock


\bibitem[\protect\citeauthoryear{Greff, Srivastava, Koutnik, Steunebrink, and
  Schmidhuber}{Greff et~al\mbox{.}}{2015}]%
        {Greff2015LSTM}
\bibfield{author}{\bibinfo{person}{K Greff}, \bibinfo{person}{R.~K.
  Srivastava}, \bibinfo{person}{J Koutnik}, \bibinfo{person}{B.~R.
  Steunebrink}, {and} \bibinfo{person}{J Schmidhuber}.}
  \bibinfo{year}{2015}\natexlab{}.
\newblock \showarticletitle{LSTM: A Search Space Odyssey}.
\newblock \bibinfo{journal}{\emph{IEEE Transactions on Neural Networks \&
  Learning Systems}} \bibinfo{volume}{28}, \bibinfo{number}{10}
  (\bibinfo{year}{2015}), \bibinfo{pages}{2222--2232}.
\newblock


\bibitem[\protect\citeauthoryear{Gregor, Danihelka, Graves, Rezende, and
  Wierstra}{Gregor et~al\mbox{.}}{2015}]%
        {Gregor2015DRAW}
\bibfield{author}{\bibinfo{person}{Karol Gregor}, \bibinfo{person}{Ivo
  Danihelka}, \bibinfo{person}{Alex Graves}, \bibinfo{person}{Danilo~Jimenez
  Rezende}, {and} \bibinfo{person}{Daan Wierstra}.}
  \bibinfo{year}{2015}\natexlab{}.
\newblock \showarticletitle{DRAW: a recurrent neural network for image
  generation}.
\newblock \bibinfo{journal}{\emph{Computer Science}} (\bibinfo{year}{2015}),
  \bibinfo{pages}{1462--1471}.
\newblock


\bibitem[\protect\citeauthoryear{Hinton, Deng, Yu, Dahl, Mohamed, Jaitly,
  Senior, Vanhoucke, Nguyen, Sainath, et~al\mbox{.}}{Hinton
  et~al\mbox{.}}{2012}]%
        {hinton2012deep}
\bibfield{author}{\bibinfo{person}{Geoffrey Hinton}, \bibinfo{person}{Li Deng},
  \bibinfo{person}{Dong Yu}, \bibinfo{person}{George~E Dahl},
  \bibinfo{person}{Abdel-rahman Mohamed}, \bibinfo{person}{Navdeep Jaitly},
  \bibinfo{person}{Andrew Senior}, \bibinfo{person}{Vincent Vanhoucke},
  \bibinfo{person}{Patrick Nguyen}, \bibinfo{person}{Tara~N Sainath},
  {et~al\mbox{.}}} \bibinfo{year}{2012}\natexlab{}.
\newblock \showarticletitle{Deep neural networks for acoustic modeling in
  speech recognition: The shared views of four research groups}.
\newblock \bibinfo{journal}{\emph{IEEE Signal processing magazine}}
  \bibinfo{volume}{29}, \bibinfo{number}{6} (\bibinfo{year}{2012}),
  \bibinfo{pages}{82--97}.
\newblock


\bibitem[\protect\citeauthoryear{Hochreiter and Schmidhuber}{Hochreiter and
  Schmidhuber}{1997}]%
        {Sepp1997Long}
\bibfield{author}{\bibinfo{person}{Sepp Hochreiter} {and}
  \bibinfo{person}{J{\"u}rgen Schmidhuber}.} \bibinfo{year}{1997}\natexlab{}.
\newblock \showarticletitle{Long short-term memory}.
\newblock \bibinfo{journal}{\emph{Neural computation}} \bibinfo{volume}{9},
  \bibinfo{number}{8} (\bibinfo{year}{1997}), \bibinfo{pages}{1735--1780}.
\newblock


\bibitem[\protect\citeauthoryear{Hudaib, Dannoun, Harfoushi, Obiedat, and
  Faris}{Hudaib et~al\mbox{.}}{2015}]%
        {Hudaib2015Hybrid}
\bibfield{author}{\bibinfo{person}{Amjad Hudaib}, \bibinfo{person}{Reham
  Dannoun}, \bibinfo{person}{Osama Harfoushi}, \bibinfo{person}{Ruba Obiedat},
  {and} \bibinfo{person}{Hossam Faris}.} \bibinfo{year}{2015}\natexlab{}.
\newblock \showarticletitle{Hybrid Data Mining Models for Predicting Customer
  Churn}.
\newblock \bibinfo{journal}{\emph{International Journal of Communications
  Network \& System Sciences}} \bibinfo{volume}{8}, \bibinfo{number}{5}
  (\bibinfo{year}{2015}), \bibinfo{pages}{91--96}.
\newblock


\bibitem[\protect\citeauthoryear{Jozefowicz, Zaremba, and Sutskever}{Jozefowicz
  et~al\mbox{.}}{2015}]%
        {Jozefowicz2015An}
\bibfield{author}{\bibinfo{person}{Rafal Jozefowicz}, \bibinfo{person}{Wojciech
  Zaremba}, {and} \bibinfo{person}{Ilya Sutskever}.}
  \bibinfo{year}{2015}\natexlab{}.
\newblock \showarticletitle{An empirical exploration of recurrent network
  architectures}. In \bibinfo{booktitle}{\emph{International Conference on
  International Conference on Machine Learning}}. \bibinfo{pages}{2342--2350}.
\newblock


\bibitem[\protect\citeauthoryear{Kalchbrenner, Danihelka, and
  Graves}{Kalchbrenner et~al\mbox{.}}{2015}]%
        {Kalchbrenner2015Grid}
\bibfield{author}{\bibinfo{person}{Nal Kalchbrenner}, \bibinfo{person}{Ivo
  Danihelka}, {and} \bibinfo{person}{Alex Graves}.}
  \bibinfo{year}{2015}\natexlab{}.
\newblock \showarticletitle{Grid Long Short-Term Memory}.
\newblock \bibinfo{journal}{\emph{Computer Science}} (\bibinfo{year}{2015}).
\newblock


\bibitem[\protect\citeauthoryear{Kasiran, Ibrahim, and Ribuan}{Kasiran
  et~al\mbox{.}}{2014}]%
        {Kasiran2014Customer}
\bibfield{author}{\bibinfo{person}{Zolidah Kasiran}, \bibinfo{person}{Zaidah
  Ibrahim}, {and} \bibinfo{person}{Muhammad Syahir~Mohd Ribuan}.}
  \bibinfo{year}{2014}\natexlab{}.
\newblock \showarticletitle{Customer churn prediction using recurrent neural
  network with reinforcement learning algorithm in mobile phone users}.
\newblock \bibinfo{journal}{\emph{International Journal of Intelligent
  Information Processing}} \bibinfo{volume}{5}, \bibinfo{number}{1}
  (\bibinfo{year}{2014}), \bibinfo{pages}{1}.
\newblock


\bibitem[\protect\citeauthoryear{Klimmt, Blake, Vorderer, and Roth}{Klimmt
  et~al\mbox{.}}{2009}]%
        {Klimmt2009Player}
\bibfield{author}{\bibinfo{person}{Christoph Klimmt},
  \bibinfo{person}{Christopher Blake}, \bibinfo{person}{Peter Vorderer}, {and}
  \bibinfo{person}{Christian Roth}.} \bibinfo{year}{2009}\natexlab{}.
\newblock \showarticletitle{Player Performance, Satisfaction, and Video Game
  Enjoyment}. In \bibinfo{booktitle}{\emph{International Conference on
  Entertainment Computing}}. \bibinfo{pages}{1--12}.
\newblock


\bibitem[\protect\citeauthoryear{Koutnik, Greff, Gomez, and
  Schmidhuber}{Koutnik et~al\mbox{.}}{2014}]%
        {Koutn2014A}
\bibfield{author}{\bibinfo{person}{Jan Koutnik}, \bibinfo{person}{Klaus Greff},
  \bibinfo{person}{Faustino Gomez}, {and} \bibinfo{person}{Juergen
  Schmidhuber}.} \bibinfo{year}{2014}\natexlab{}.
\newblock \showarticletitle{A Clockwork RNN}.
\newblock \bibinfo{journal}{\emph{Computer Science}} (\bibinfo{year}{2014}),
  \bibinfo{pages}{1863--1871}.
\newblock


\bibitem[\protect\citeauthoryear{Lazzaro}{Lazzaro}{2004}]%
        {Lazzaro2004Why}
\bibfield{author}{\bibinfo{person}{Nicole Lazzaro}.}
  \bibinfo{year}{2004}\natexlab{}.
\newblock \showarticletitle{Why we play games: Four keys to more emotion
  without story}.
\newblock  (\bibinfo{year}{2004}).
\newblock


\bibitem[\protect\citeauthoryear{Liaw, Wiener, et~al\mbox{.}}{Liaw
  et~al\mbox{.}}{2002}]%
        {liaw2002classification}
\bibfield{author}{\bibinfo{person}{Andy Liaw}, \bibinfo{person}{Matthew
  Wiener}, {et~al\mbox{.}}} \bibinfo{year}{2002}\natexlab{}.
\newblock \showarticletitle{Classification and regression by randomForest}.
\newblock \bibinfo{journal}{\emph{R news}} \bibinfo{volume}{2},
  \bibinfo{number}{3} (\bibinfo{year}{2002}), \bibinfo{pages}{18--22}.
\newblock


\bibitem[\protect\citeauthoryear{Phan, Keebler, and Chaparro}{Phan
  et~al\mbox{.}}{2016}]%
        {Phan2016The}
\bibfield{author}{\bibinfo{person}{Mikki~H Phan}, \bibinfo{person}{Joseph~R
  Keebler}, {and} \bibinfo{person}{Barbara~S Chaparro}.}
  \bibinfo{year}{2016}\natexlab{}.
\newblock \showarticletitle{The development and validation of the game user
  experience satisfaction scale (GUESS)}.
\newblock \bibinfo{journal}{\emph{Human factors}} \bibinfo{volume}{58},
  \bibinfo{number}{8} (\bibinfo{year}{2016}), \bibinfo{pages}{1217--1247}.
\newblock


\bibitem[\protect\citeauthoryear{Wang, Yin, Deng, Li, Pu, Tang, and Luo}{Wang
  et~al\mbox{.}}{2018}]%
        {Wang2018Evaluating}
\bibfield{author}{\bibinfo{person}{Junxiang Wang}, \bibinfo{person}{Jianwei
  Yin}, \bibinfo{person}{Shuiguang Deng}, \bibinfo{person}{Ying Li},
  \bibinfo{person}{Calton Pu}, \bibinfo{person}{Yan Tang}, {and}
  \bibinfo{person}{Zhiling Luo}.} \bibinfo{year}{2018}\natexlab{}.
\newblock \showarticletitle{Evaluating User Satisfaction with Typography
  Designs via Mining Touch Interaction Data in Mobile Reading}. In
  \bibinfo{booktitle}{\emph{CHI Conference}}. \bibinfo{pages}{1--12}.
\newblock


\bibitem[\protect\citeauthoryear{Yannakakis}{Yannakakis}{2008}]%
        {Yannakakis2008How}
\bibfield{author}{\bibinfo{person}{Georgios~N Yannakakis}.}
  \bibinfo{year}{2008}\natexlab{}.
\newblock \showarticletitle{How to model and augment player satisfaction: a
  review}. In \bibinfo{booktitle}{\emph{First Workshop on Child, Computer and
  Interaction}}.
\newblock


\end{thebibliography}

\end{document}